\documentclass[10pt,twocolumn,letterpaper]{article}

\usepackage{iccv}
\usepackage{times}
\usepackage{epsfig}
\usepackage{graphicx}
\usepackage{amsmath}
\usepackage{amssymb}
\usepackage{xfrac}
\usepackage{subcaption}
\newcommand{\norm}[1]{\left\lVert #1 \right\rVert}
\usepackage{enumitem}

\usepackage[pagebackref=true,breaklinks=true,letterpaper=true,colorlinks,bookmarks=false]{hyperref}

\iccvfinalcopy 

\def\iccvPaperID{1007} 

\begin{document}

\title{End-to-End Learning of Geometry and Context for Deep Stereo Regression}

\author{Alex Kendall \qquad Hayk Martirosyan \qquad Saumitro Dasgupta \qquad Peter Henry\\
Ryan Kennedy \qquad Abraham Bachrach \qquad Adam Bry\vspace{0.5em}\\
\large Skydio Inc.\\
\normalsize
\texttt{\{alex,hayk,saumitro,peter,ryan,abe,adam\}@skydio.com}
}

\newcommand{\fig}[1]{Figure~\ref{fig:#1}}
\newcommand{\tbl}[1]{Table~\ref{tbl:#1}}

\maketitle

\begin{abstract}

We propose a novel deep learning architecture for regressing disparity from a rectified pair of stereo images. We leverage knowledge of the problem's geometry to form a cost volume using deep feature representations. We learn to incorporate contextual information using {3-D} convolutions over this volume. 
Disparity values are regressed from the cost volume using a proposed differentiable soft argmin operation, which allows us to train our method end-to-end to sub-pixel accuracy without any additional post-processing or regularization. We evaluate our method on the Scene Flow and KITTI datasets and on KITTI we set a new state-of-the-art benchmark, while being significantly faster than competing approaches.

\end{abstract}

\section{Introduction}

\begin{figure*}[t]
	\begin{center}
		\includegraphics[width=\linewidth]{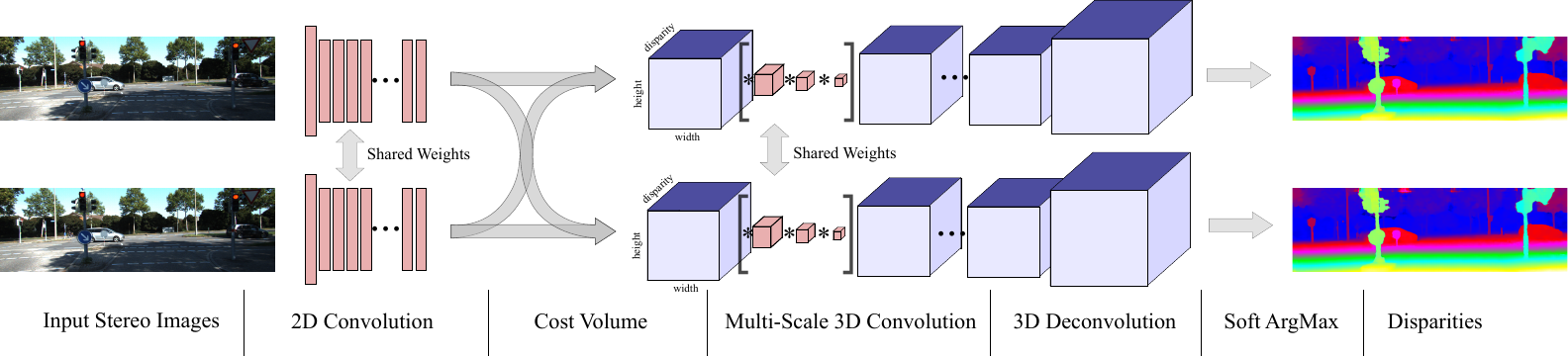}
	\end{center}
	\caption{\textbf{Our end-to-end deep stereo regression architecture, GC-Net} (\underline{G}eometry and \underline{C}ontext \underline{Net}work).}
	\label{fig:model}
\end{figure*}

Accurately estimating three dimensional geometry from stereo imagery is a core problem for many computer vision applications, including autonomous vehicles and UAVs \cite{achtelik2009stereo}. In this paper we are specifically interested in computing the disparity of each pixel between a rectified stereo pair of images. To achieve this, the core task of a stereo algorithm is computing the correspondence of each pixel between two images. This is very challenging to achieve robustly in real-world scenarios. Current state-of-the-art stereo algorithms often have difficulty with textureless areas, reflective surfaces, thin structures and repetitive patterns. Many stereo algorithms aim to mitigate these failures with pooling or gradient based regularization \cite{geiger2010efficient,hirschmuller2005accurate}. However, this often requires a compromise between smoothing surfaces and detecting detailed structures.

In contrast, deep learning models have been successful in learning powerful representations directly from the raw data in object classification \cite{krizhevsky2012imagenet}, detection \cite{girshick2014rich} and semantic segmentation \cite{long2015fully,badrinarayanan2015segnet}. These examples demonstrate that deep convolutional neural networks are very effective for understanding semantics. They excel at classification tasks when supervised with large training datasets. We observe that a number of these challenging problems for stereo algorithms would benefit from knowledge of global semantic context, rather than relying solely on local geometry. For example, given a reflective surface of a vehicle's wind-shield, a stereo algorithm is likely to be erroneous if it relies solely on the local appearance of the reflective surface to compute geometry. Rather, it would be advantageous to understand the semantic context of this surface (that it belongs to a vehicle) to infer the local geometry. In this paper we show how to learn a stereo regression model which can be trained end-to-end, with the capacity to understand wider contextual information.

Stereo algorithms which leverage deep learning representations have so far been largely focused on using them to generate unary terms \cite{zbontar2015computing,luo2016efficient}. Applying cost matching on the deep unary representations performs poorly when estimating pixel disparities \cite{luo2016efficient,zbontar2015computing}. Traditional regularization and post processing steps are still used, such as semi global block matching and left-right consistency checks \cite{hirschmuller2005accurate}. These regularization steps are severely limited because they are hand-engineered, shallow functions, which are still susceptible to the aforementioned problems.

This paper asks the question, can we formulate the entire stereo vision problem with deep learning using our understanding of stereo geometry? The main contribution of this paper is an end-to-end deep learning method to estimate per-pixel disparity from a single rectified image pair. Our architecture is illustrated in \fig{model}. It explicitly reasons about geometry by forming a cost volume, while also reasoning about semantics using a deep convolutional network formulation. We achieve this with two key ideas:
\begin{itemize}[topsep=0pt,itemsep=-1ex,partopsep=1ex,parsep=1ex]
\item We learn to incorporate context directly from the data, employing 3-D convolutions to learn to regularize the cost volume over $height\times width\times disparity$ dimensions,
\item We use a soft argmin function, which is fully differentiable, and allows us to regress sub-pixel disparity values from the disparity cost volume.
\end{itemize}

Section \ref{section:model} introduces this model and illustrates these components in more detail. In Section \ref{section:evaluation} we evaluate our model on the synthetic Scene Flow dataset \cite{MIFDB16} and set a new state-of-the-art benchmark on the KITTI 2012 and 2015 datasets \cite{Geiger2012CVPR,Menze2015CVPR}. Finally, in Section \ref{sec:saliency} we present evidence that our model has the capacity to learn semantic reasoning and contextual information.

\section{Related Work}
The problem of computing depth from stereo image pairs has been studied for quite some time~\cite{Barnard1982}. 
A survey by Scharstein and Szeliski~\cite{Scharstein2002} provides a taxonomy of stereo algorithms as performing some subset of: matching cost computation, cost support aggregation, disparity computation and optimization, or disparity refinement. This survey also described the first Middlebury dataset and associated evaluation metrics, using structured light to provide ground truth.  
The KITTI dataset~\cite{Geiger2012CVPR,Menze2015CVPR} is a larger dataset from data collected from a moving vehicle with ground truth supplied by LIDAR. These datasets first motivated improved hand-engineered techniques for all components of stereo, of which we mention a few notable examples.  

The matching cost is a measure of pixel dissimilarity for potentially corresponding image locations \cite{Hirschmuller2007}, of which absolute differences, squared differences, and truncated differences are examples.  
Local descriptors based on gradients~\cite{Geiger2010} or binary patterns, such as CENSUS~\cite{Zabih1994} or BRIEF~\cite{Calonder2010,Heise2015}, can be employed.  
Instead of aggregating neighboring pixels equally as patch-based matching costs do, awareness of the image content can more heavily incorporate neighboring pixels possessing similar appearance, under the assumption that they are more likely to come from the same surface and disparity.  A survey of these techniques is provided by Tombari et al.~\cite{Tombari2008}.  
Local matching costs may also be optimized within a global framework, usually minimizing an energy function combining a local data term and a pairwise smoothness term.  Global optimization can be accomplished using graph cuts~\cite{Kolmogorov2001} or belief propagation~\cite{Klaus2006}, which can be extended to slanted surfaces~\cite{Bleyer2011}. A popular and effective approximation to global optimization is the \emph{Semi-Global Matching} (SGM) of Hirschm{\"u}ller~\cite{Hirschmuller2008}, where dynamic programming optimizes a pathwise form of the energy function in many directions.

In addition to providing a basis for comparing stereo algorithms, the ground truth depth data from these datasets provides the opportunity to use machine learning for improving stereo algorithms in a variety of ways.  Zhang and Seitz~\cite{Zhang2007} alternately optimized disparity and Markov random field regularization parameters.  Scharstein and Pal~\cite{Scharstein2007} learn conditional random field (CRF) parameters, and Li and Huttenlocher~\cite{Li2008} train a non-parametric CRF model using the structured support vector machine.  Learning can also be employed to estimate the confidence of a traditional stereo algorithm, such as the random forest approach of Haeusler et al.~\cite{Haeusler2013a}.  Such confidence measures can improve the result of SGM as shown by Park and Yoon~\cite{Park2015}.

Deep convolutional neural networks can be trained to match image patches~\cite{Zagoruyko2015}.  A deep network trained to match $9 \times 9$ image patches, followed by non-learned cost aggregation and regularization, was shown by {\v Z}bontar and LeCun~\cite{Zbontar2015a,Zbontar2015} to produce then state-of-the-art results.  Luo et al. presented a notably faster network for computing local matching costs as a multi-label classification of disparities using a Siamese network~\cite{Luo2016}.  A multi-scale embedding model from Chen et al.~\cite{Chen2016} also provided good local matching scores.  Also noteworthy is the \emph{DeepStereo} work of Flynn et al.~\cite{Flynn2016}, which learns a cost volume combined with a separate conditional color model to predict novel viewpoints in a multi-view stereo setting.

Mayer et al. created a large synthetic dataset to train a network for disparity estimation (as well as optical flow)~\cite{Mayer2015}, improving the state-of-the-art. As one variant of the network, a 1-D correlation was proposed along the disparity line which is a multiplicative approximation to the stereo cost volume. In addition, this volume is concatenated with convolutional features from a single image and succeeded by a series of further convolutions. In contrast, our work does not collapse the feature dimension when computing the cost volume and uses 3-D convolutions to incorporate context.

Though the focus of this work is on binocular stereo, it is worth noting that the representational power of deep convolutional networks also enables depth estimation from a single monocular image~\cite{Eigen2014a}.  Deep learning is combined with a continuous CRF by Liu et al.~\cite{Liu2015}.  Instead of supervising training with labeled ground truth, unlabeled stereo pairs can be used to train a monocular model~\cite{Garg2016}.

In our work, we apply no post-processing or regularization. Our network can explicitly reason about geometry by forming a fully differentiable cost volume. Our network learns to incorporate context from the data with a 3-D convolutional architecture.  We don't learn a probability distribution, cost function, or classification result.  Rather, our network is able to directly regress a sub-pixel estimate of disparity from a stereo image pair.

\section{Learning End-to-end Disparity Regression}
\label{section:model}

Rather than design any step of the stereo algorithm by hand, we would like to learn an end-to-end mapping from an image pair to disparity maps using deep learning. We hope to learn a more optimal function directly from the data. Additionally, this approach promises to reduce much of the engineering design complexity. However, our intention is not to naively construct a machine learning architecture as a black box to model stereo. Instead, we advocate the use of the insights from many decades of multi-view geometry research \cite{hartley2003multiple} to guide architectural design. Therefore, we form our model by developing differentiable layers representing each major component in traditional stereo pipelines \cite{Scharstein2002}. This allows us to learn the entire model end-to-end while leveraging our geometric knowledge of the stereo problem. 

Our architecture, GC-Net (\underline{G}eometry and \underline{C}ontext \underline{Net}work) is illustrated in \fig{model}, with a more detailed layer-by-layer definition in \tbl{model}. In the remainder of this section we discuss each component in detail. Later, in Section \ref{sec:model_results}, we present quantitative results justifying our design decisions.

\begin{table}[t]
\centering
\resizebox{\linewidth}{!}{
\begin{tabular}{l|l|c}
& Layer Description & Output Tensor Dim. \\ \hline \hline
& Input image & H$\times$W$\times$C \\ \hline
\multicolumn{3}{c}{\textbf{Unary features (section \ref{sec:unary})}} \\ \hline
1 & 5$\times$5 conv, 32 features, stride 2 & \sfrac{1}{2}H$\times$\sfrac{1}{2}W$\times$F \\
2 & 3$\times$3 conv, 32 features & \sfrac{1}{2}H$\times$\sfrac{1}{4}W$\times$F \\
3 & 3$\times$3 conv, 32 features & \sfrac{1}{2}H$\times$\sfrac{1}{4}W$\times$F \\
 & add layer 1 and 3 features (residual connection) & \sfrac{1}{2}H$\times$\sfrac{1}{2}W$\times$F \\
4-17 & (repeat layers 2,3 and residual connection) $\times$ 7 & \sfrac{1}{2}H$\times$\sfrac{1}{2}W$\times$F \\
18 & 3$\times$3 conv, 32 features, (no ReLu or BN) & \sfrac{1}{2}H$\times$\sfrac{1}{2}W$\times$F \\ \hline
\multicolumn{3}{c}{\textbf{Cost volume (section \ref{sec:cost_vol})}} \\ \hline
 & Cost Volume & \sfrac{1}{2}D$\times$\sfrac{1}{2}H$\times$\sfrac{1}{2}W$\times$2F \\ \hline
\multicolumn{3}{c}{\textbf{Learning regularization (section \ref{sec:regularise})}} \\ \hline
19 & 3-D conv, 3$\times$3$\times$3, 32 features & \sfrac{1}{2}D$\times$\sfrac{1}{2}H$\times$\sfrac{1}{2}W$\times$F \\
20 & 3-D conv, 3$\times$3$\times$3, 32 features & \sfrac{1}{2}D$\times$\sfrac{1}{2}H$\times$\sfrac{1}{2}W$\times$F \\
%
21 & From 18: 3-D conv, 3$\times$3$\times$3, 64 features, stride 2 & \sfrac{1}{4}D$\times$\sfrac{1}{4}H$\times$\sfrac{1}{4}W$\times$2F \\
22 & 3-D conv, 3$\times$3$\times$3, 64 features & \sfrac{1}{4}D$\times$\sfrac{1}{4}H$\times$\sfrac{1}{4}W$\times$2F \\
23 & 3-D conv, 3$\times$3$\times$3, 64 features & \sfrac{1}{4}D$\times$\sfrac{1}{4}H$\times$\sfrac{1}{4}W$\times$2F \\
%
24 & From 21: 3-D conv, 3$\times$3$\times$3, 64 features, stride 2 & \sfrac{1}{8}D$\times$\sfrac{1}{8}H$\times$\sfrac{1}{8}W$\times$2F \\
25 & 3-D conv, 3$\times$3$\times$3, 64 features & \sfrac{1}{8}D$\times$\sfrac{1}{8}H$\times$\sfrac{1}{8}W$\times$2F \\
26 & 3-D conv, 3$\times$3$\times$3, 64 features & \sfrac{1}{8}D$\times$\sfrac{1}{8}H$\times$\sfrac{1}{8}W$\times$2F \\
%
27 & From 24: 3-D conv, 3$\times$3$\times$3, 64 features, stride 2 & \sfrac{1}{16}D$\times$\sfrac{1}{16}H$\times$\sfrac{1}{16}W$\times$2F\\
28 & 3-D conv, 3$\times$3$\times$3, 64 features & \sfrac{1}{16}D$\times$\sfrac{1}{16}H$\times$\sfrac{1}{16}W$\times$2F \\
29 & 3-D conv, 3$\times$3$\times$3, 64 features & \sfrac{1}{16}D$\times$\sfrac{1}{16}H$\times$\sfrac{1}{16}W$\times$2F \\
%
30 & From 27: 3-D conv, 3$\times$3$\times$3, 128 features, stride 2 & \sfrac{1}{32}D$\times$\sfrac{1}{32}H$\times$\sfrac{1}{32}W$\times$4F\\
31 & 3-D conv, 3$\times$3$\times$3, 128 features & \sfrac{1}{32}D$\times$\sfrac{1}{32}H$\times$\sfrac{1}{32}W$\times$4F \\
32 & 3-D conv, 3$\times$3$\times$3, 128 features & \sfrac{1}{32}D$\times$\sfrac{1}{32}H$\times$\sfrac{1}{32}W$\times$4F \\
%
33 & 3$\times$3$\times$3, 3-D transposed conv, 64 features, stride 2 & \sfrac{1}{16}D$\times$\sfrac{1}{16}H$\times$\sfrac{1}{16}W$\times$2F \\
 & add layer 33 and 29 features (residual connection) & \sfrac{1}{16}D$\times$\sfrac{1}{16}H$\times$\sfrac{1}{16}W$\times$2F \\
34 & 3$\times$3$\times$3, 3-D transposed conv, 64 features, stride 2 & \sfrac{1}{8}D$\times$\sfrac{1}{8}H$\times$\sfrac{1}{8}W$\times$2F \\
 & add layer 34 and 26 features (residual connection) & \sfrac{1}{8}D$\times$\sfrac{1}{8}H$\times$\sfrac{1}{8}W$\times$2F \\
35 & 3$\times$3$\times$3, 3-D transposed conv, 64 features, stride 2 & \sfrac{1}{4}D$\times$\sfrac{1}{4}H$\times$\sfrac{1}{4}W$\times$2F \\
 & add layer 35 and 23 features (residual connection) & \sfrac{1}{4}D$\times$\sfrac{1}{4}H$\times$\sfrac{1}{4}W$\times$2F \\
36 & 3$\times$3$\times$3, 3-D transposed conv, 32 features, stride 2 & \sfrac{1}{2}D$\times$\sfrac{1}{2}H$\times$\sfrac{1}{2}W$\times$F \\
 & add layer 36 and 20 features (residual connection) & \sfrac{1}{2}D$\times$\sfrac{1}{2}H$\times$\sfrac{1}{2}W$\times$F \\
%
37 & 3$\times$3$\times$3, 3-D trans conv, 1 feature (no ReLu or BN) & D$\times$H$\times$W$\times$1 \\ \hline
\multicolumn{3}{c}{\textbf{Soft argmin (section \ref{sec:argmin})}} \\ \hline
 & Soft argmin & H$\times$W \\
\end{tabular}}
	\caption{Summary of our end-to-end deep stereo regression architecture, GC-Net. Each 2-D or 3-D convolutional layer represents a block of convolution, batch normalization and ReLU non-linearity (unless otherwise specified).}
	\label{tbl:model}
\end{table}

\subsection{Unary Features}
\label{sec:unary}

First we learn a deep representation to use to compute the stereo matching cost. Rather than compute the stereo matching cost using raw pixel intensities, it is common to use a feature representation. The motivation is to compare a descriptor which is more robust to the ambiguities in photometric appearance and can incorporate local context.

In our model we learn a deep representation through a number of 2-D convolutional operations. Each convolutional layer is followed by a batch normalization layer and a rectified linear non-linearity. To reduce computational demand, we initially apply a 5$\times$5 convolutional filter with stride of two to subsample the input. Following this layer, we append eight residual blocks \cite{he2015deep} which each consist of two 3$\times$3 convolutional filters in series. Our final model architecture is shown in \tbl{model}. We form the unary features by passing both left and right stereo images through these layers. We share the parameters between the left and right towers to more effectively learn corresponding features.

\subsection{Cost Volume}
\label{sec:cost_vol}

We use the deep unary features to compute the stereo matching cost by forming a cost volume. While a naive approach might simply concatenate the left and right feature maps, forming a cost volume allows us to constrain the model in a way which preserves our knowledge of the geometry of stereo vision. For each stereo image, we form a cost volume of dimensionality {\it height}$\times${\it width}$\times${(\it max disparity + 1)}$\times${\it feature size}. We achieve this by concatenating each unary feature with their corresponding unary from the opposite stereo image across each disparity level, and packing these into the 4D volume. 

Crucially, we retain the feature dimension through this operation, unlike previous work which uses a dot product style operation which decimates the feature dimension \cite{luo2016efficient}. This allows us to learn to incorporate context which can operate over feature unaries (Section \ref{sec:regularise}). We find that forming a cost volume with concatenated features improves performance over subtracting features or using a distance metric. Our intuition is that by maintaining the feature unaries, the network has the opportunity to learn an absolute representation (because it is not a distance metric) and carry this through to the cost volume. This gives the architecture the capacity to learn semantics. In contrast, using a distance metric restricts the network to only learning relative representations between features, and cannot carry absolute feature representations through to cost volume.

\begin{figure*}[t]
	\begin{center}
    		\begin{subfigure}[b]{0.32\linewidth}
			\includegraphics[width=\linewidth]{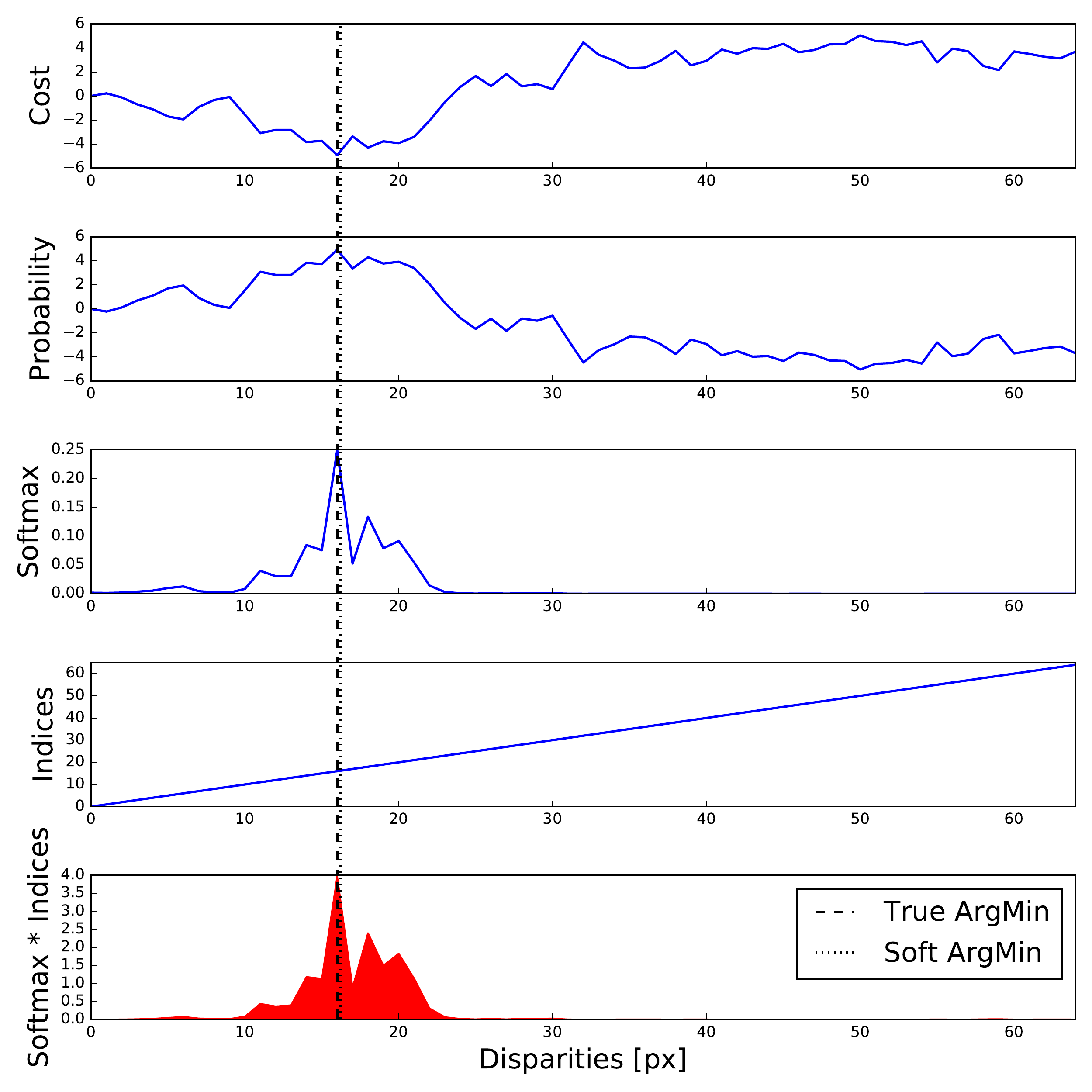}
	        \caption{Soft ArgMin}
		\end{subfigure}
    		\begin{subfigure}[b]{0.32\linewidth}
			\includegraphics[width=\linewidth]{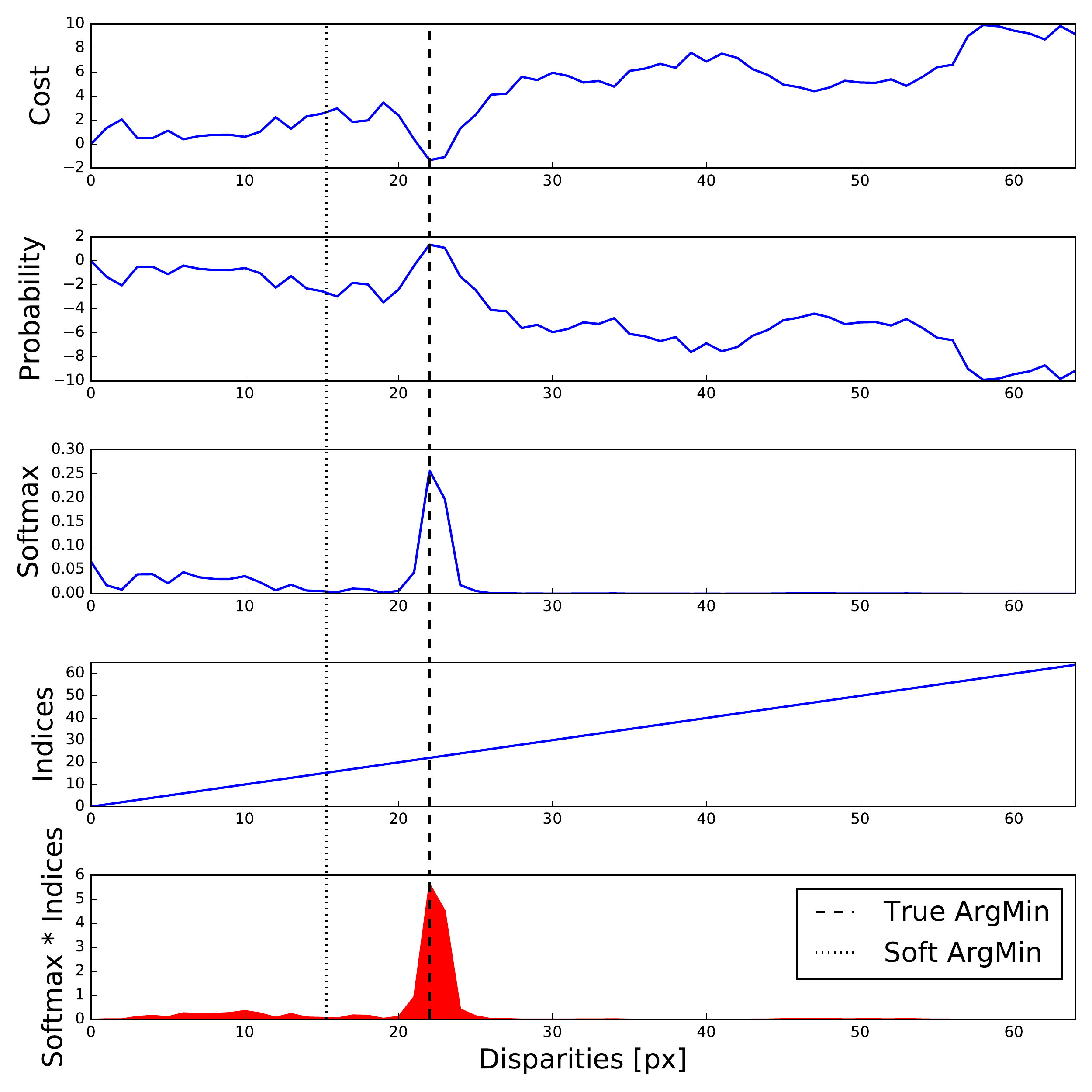}
	        \caption{Multi-modal distribution}
		\end{subfigure}
    		\begin{subfigure}[b]{0.32\linewidth}
			\includegraphics[width=\linewidth]{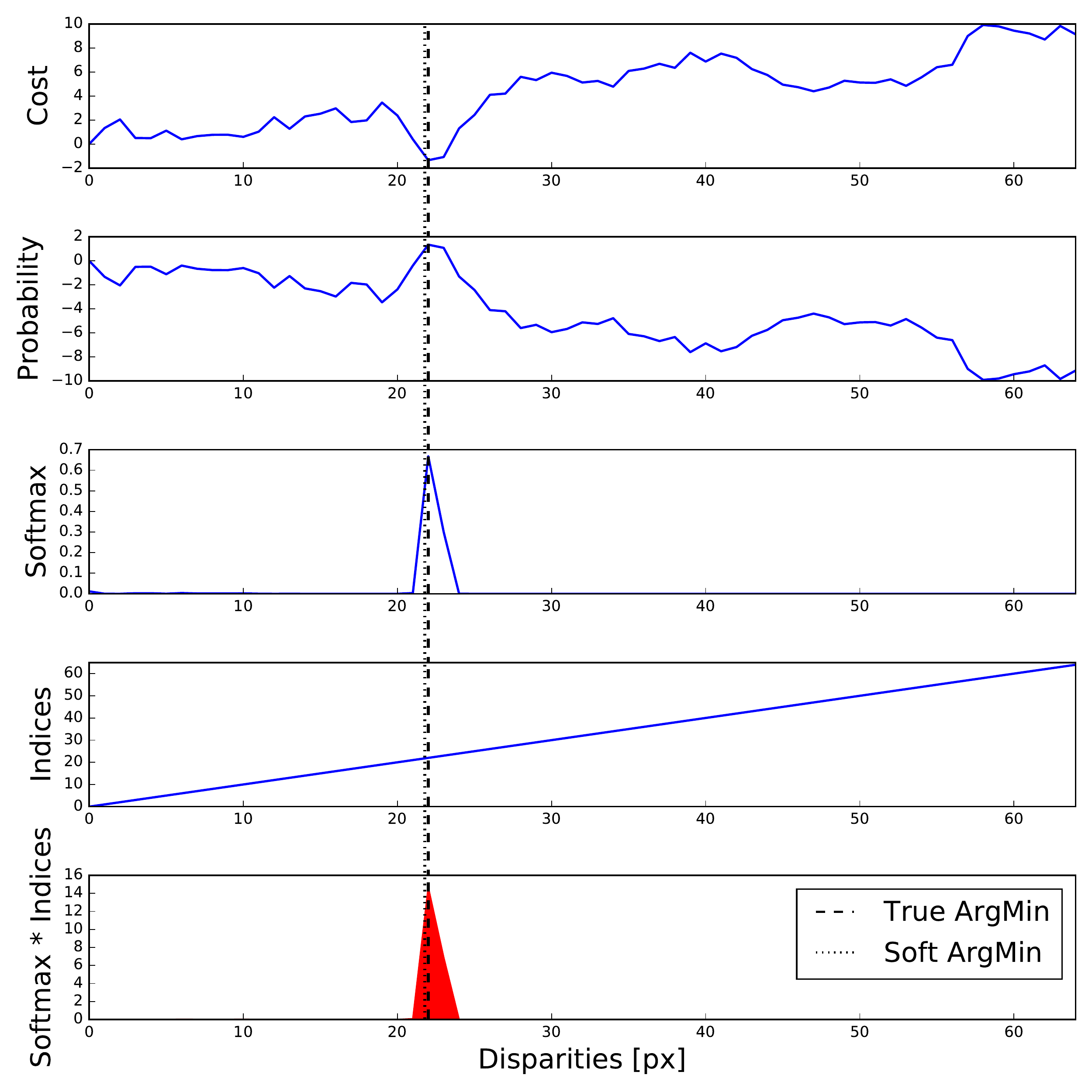}
	        \caption{Multi-modal distribution with prescaling}
		\end{subfigure}
	\end{center}
	\caption{\textbf{A graphical depiction of the soft argmin operation} (Section \ref{sec:argmin}) which we propose in this work. It is able to take a cost curve along each disparity line and output an estimate of the argmin by summing the product of each disparity's softmax probability and it's disparity index. (a) demonstrates that this very accurately captures the true argmin when the curve is uni-modal. (b) demonstrates a failure case when the data is bi-modal with one peak and one flat region. (c) demonstrates that this failure may be avoided if the network learns to pre-scale the cost curve, because the softmax probabilities will tend to be more extreme, producing a uni-modal result.}
	\label{fig:argmin}
\end{figure*}

\subsection{Learning Context}
\label{sec:regularise}

Given this disparity cost volume, we would now like to learn a regularization function which is able to take into account context in this volume and refine our disparity estimate. The matching costs between unaries can never be perfect, even when using a deep feature representation. For example, in regions of uniform pixel intensity (for example, sky) the cost curve will be flat for any features based on a fixed, local context.  We find that regions like this can cause multi modal matching cost curves across the disparity dimension. Therefore we wish to learn to regularize and improve this volume.

We propose to use three-dimensional convolutional operations to filter and refine this representation. 3-D convolutions are able to learn feature representations from the height, width and disparity dimensions. Because we compute the cost curve for each unary feature, we can learn convolutional filters from this representation. In Section \ref{sec:model_results} we show the importance of these 3-D filters for learning context and significantly improving stereo performance.


The difficulty with 3-D convolutions is that the additional dimension is a burden on the computational time for both training and inference. Deep encoder-decoder tasks which are designed for dense prediction tasks get around their computational burden by encoding sub-sampled feature maps, followed by up-sampling in a decoder \cite{badrinarayanan2015segnet}. We extend this idea to three dimensions. By sub-sampling the input with stride two, we also reduce the 3-D cost volume size by a factor of eight. We form our 3-D regularization network with four levels of sub-sampling. As the unaries are already sub-sampled by a factor of two, the features are sub-sampled by a total factor of 32. This allows us to explicitly leverage context with a wide field of view. We apply two 3$\times$3$\times$3 convolutions in series for each encoder level. To make dense predictions with the original input resolution, we employ a 3-D transposed convolution to up-sample the volume in the decoder. The full architecture is described in \tbl{model}.

Sub-sampling is useful to increase each feature's receptive field while reducing computation. However, it also reduces spatial accuracy and fine-grained details through the loss of resolution. For this reason, we add each higher resolution feature map before up-sampling. These residual layers have the benefit of retaining higher frequency information, while the up-sampled features provide an attentive feature map with a larger field of view.

Finally, we apply a single 3-D transposed convolution (deconvolution), with stride two and a single feature output. This layer is necessary to make dense prediction in the original input dimensions because the feature unaries were sub-sampled by a factor of two. This results in the final, regularized cost volume with size H$\times$W$\times$D.

\subsection{Differentiable ArgMin}
\label{sec:argmin}

Typically, stereo algorithms produce a final cost volume from the matching cost unaries. From this volume, we may estimate disparity by performing an argmin operation over the cost volume’s disparity dimension. However, this operation has two problems:
\begin{itemize}[noitemsep]
\item it is discrete and is unable to produce sub-pixel disparity estimates,
\item it is not differentiable and therefore unable to be trained using back-propagation.
\end{itemize}
To overcome these limitations, we define a \textit{soft argmin}\footnote{Note that if we wished for our network to learn probabilities, rather than cost, this function could easily be adapted to a soft argmax operation.} which is both fully differentiable and able to regress a smooth disparity estimate. First, we convert the predicted costs, $c_d$ (for each disparity, $d$) from the cost volume to a probability volume by taking the negative of each value. We normalize the probability volume across the disparity dimension with the softmax operation, $\sigma (\cdot)$. We then take the sum of each disparity, $d$, weighted by its normalized probability. A graphical illustration is shown in \fig{argmin} and defined mathematically in (\ref{eqn:argmin}):
\begin{equation}
soft~argmin := \sum_{d=0}^{D_{max}} d \times \sigma (-c_d)
\label{eqn:argmin}
\end{equation}
This operation is fully differentiable and allows us to train and regress disparity estimates. We note that a similar function was first introduced by \cite{bahdanau2014neural} and referred to as a soft-attention mechanism. Here, we show how to apply it for the stereo regression problem.

However, compared to the argmin operation, its output is influenced by all values. This leaves it susceptible to multi-modal distributions, as the output will not take the most likely. Rather, it will estimate a weighted average of all modes. To overcome this limitation, we rely on the network's regularization to produce a disparity probability distribution which is predominantly unimodal. The network can also pre-scale the matching costs to control the peakiness (sometimes called temperature) of the normalized post-softmax probabilities (\fig{argmin}). We explicitly omit batch normalization from the final convolution layer in the unary tower to allow the network to learn this from the data.

\subsection{Loss}

We train our entire model end-to-end from a random initialization. We train our model with supervised learning using ground truth depth data. In the case of using LIDAR to label ground truth values (e.g. KITTI dataset \cite{Geiger2012CVPR,Menze2015CVPR}) these labels may be sparse. Therefore, we average our loss over the labeled pixels, $N$. We train our model using the absolute error between the ground truth disparity, $d_n$, and the model's predicted disparity, $\hat{d}_n$, for pixel $n$. This supervised regression loss is defined in (\ref{eqn:loss}):
\begin{equation}
\label{eqn:loss}
Loss = \frac{1}{N} \sum\limits_{n=1}^N \norm{{d_n}-\hat{d_n}}_1
\end{equation}
In the following section we show that formulating our model as a regression problem allows us to regress with sub-pixel accuracy and outperform classification approaches. Additionally, formulating a regression model makes it possible to leverage unsupervised learning losses based on photometric reprojection error \cite{Garg2016}.


%
%
%

\begin{table*}[t]
\centering
\resizebox{\linewidth}{!}{
\begin{tabular}{l|c|c|c|c|c|c|c}
Model & $>1$ px & $>3$ px & $>5$ px & MAE (px) & RMS (px) & Param. & Time (ms) \\ \hline \hline
\multicolumn{8}{c}{\textit{1. Comparison of architectures}} \\ \hline
Unaries only (omitting all 3-D conv layers 19-36) w Regression Loss & 97.9 & 93.7 & 89.4 & 36.6 & 47.6 & 0.16M & 0.29 \\
Unaries only (omitting all 3-D conv layers 19-36) w Classification Loss & 51.9 & 24.3 & 21.7 & 13.1 & 36.0 & 0.16M & 0.29 \\
Single scale 3-D context (omitting 3-D conv layers 21-36) & 34.6 & 24.2 & 21.2 & 7.27 & 20.4 & 0.24M & 0.84 \\
\textbf{Hierarchical 3-D context (all 3-D conv layers)} & 16.9 & 9.34 & 7.22 & 2.51 & 12.4 & 3.5M & 0.95 \\ \hline
\multicolumn{8}{c}{\textit{2. Comparison of loss functions}} \\ \hline
GC-Net + Classification loss & 19.2 & 12.2 & 10.4 & 5.01 & 20.3 & 3.5M & 0.95 \\
GC-Net + Soft classification loss \cite{luo2016efficient} & 20.6 & 12.3 & 10.4 & 5.40 & 25.1 & 3.5M & 0.95 \\
\textbf{GC-Net + Regression loss} & 16.9 & 9.34 & 7.22 & 2.51 & 12.4 & 3.5M & 0.95 \\ \hline \hline
\textbf{GC-Net (final architecture with regression loss)} & 16.9 & 9.34 & 7.22 & 2.51 & 12.4 & 3.5M & 0.95 \\
\end{tabular}}
	\caption{\textbf{Results on the Scene Flow dataset} \cite{MIFDB16} which contains $35,454$ training and $4,370$ testing images of size $960\times540$px from an array of synthetic scenes. We compare different architecture variants to justify our design choices. The first experiment shows the importance of the 3-D convolutional architecture. The second experiment shows the gain in performance we get from using a regression loss.
    }
	\label{tbl:scene_flow}
\end{table*}

\section{Experimental Evaluation}
\label{section:evaluation}

In this section we present qualitative and quantitative results on two datasets, Scene Flow \cite{MIFDB16} and KITTI \cite{Geiger2012CVPR,Menze2015CVPR}. Firstly, in Section \ref{sec:model_results} we experiment with different variants of our model and justify a number of our design choices using the Scene Flow dataset \cite{MIFDB16}. In Section \ref{sec:kitti_results} we present results of our approach on the KITTI dataset and set a new state-of-the-art benchmark. Finally, we measure our model's capacity to learn context in Section \ref{sec:saliency}.

For the experiments in this section, we implement our architecture using TensorFlow \cite{tensorflow2015-whitepaper}. All models are optimized end-to-end with RMSProp \cite{tieleman2012lecture} and a constant learning rate of $1 \times 10^{-3}$. We train with a batch size of 1 using a $256\times512$ randomly located crop from the input images. Before training we normalize each image such that the pixel intensities range from $-1$ to $1$. We trained the network (from a random initialization) on Scene Flow for approximately 150k iterations which takes two days on a single NVIDIA Titan-X GPU. For the KITTI dataset we fine-tune the models pre-trained on Scene Flow for a further 50k iterations. For our experiments on Scene Flow we use F=32, H=540, W=960, D=192 and on the KITTI dataset we use F=32, H=388, W=1240, D=192 for feature size, image height, image width and maximum disparity, respectively.

\subsection{Model Design Analysis}
\label{sec:model_results}

In \tbl{scene_flow} we present an ablation study to compare a number of different model variants and justify our design choices. We wish to evaluate the importance of the key ideas in this paper; using a regression loss over a classification loss, and learning 3-D convolutional filters for cost volume regularization. We use the synthetic Scene Flow dataset \cite{MIFDB16} for these experiments, which contains $35,454$ training and $4,370$ testing images. We use this dataset for two reasons. Firstly, we know perfect, dense ground truth from the synthetic scenes which removes any discrepancies due to erroneous labels. Secondly, the dataset is large enough to train the model without over-fitting. In contrast, the KITTI dataset only contains 200 training images, and we observe that the model is susceptible to over-fitting to this very small dataset. With tens of thousands of training images we do not have to consider over-fitting in our evaluation.

The first experiment in \tbl{scene_flow} shows that including the 3-D filters performs significantly better than learning unaries only. We compare our full model (as defined in \tbl{model}) to a model which uses only unary features (omitting all 3-D convolutional layers 19-36) and a model which omits the hierarchical 3-D convolution (omitting layers 21-36).  We observe that the 3-D filters are able to regularize and smooth the output effectively, while learning to retain sharpness and accuracy in the output disparity map. We find that the hierarchical 3-D model outperforms the vanilla 3-D convolutional model by aggregating a much large context, without significantly increasing computational demand.

The second experiment in \tbl{scene_flow} compares our regression loss function to baselines which classify disparities using hard or soft classification as proposed in \cite{luo2016efficient}. Hard classification trains the network to classify disparities in the cost volume as probabilities using cross entropy loss with a `one hot' encoding. Soft classification (used by \cite{luo2016efficient}) smooths this encoding to learn a Gaussian distribution centered around the correct disparity value. In \tbl{scene_flow} we observe that our regression approach outperforms both hard and soft classification. This is especially noticeable for the pixel accuracy metrics and the percentage of pixels which are within one pixel of the true disparity, because the regression loss allows the model to predict with sub-pixel accuracy.

\begin{table*}[t]
\centering
\begin{subtable}[t]{\linewidth}
\centering
\begin{tabular}{l|cc|cc|cc|cc|c} \hline
& \multicolumn{2}{c|}{\textgreater 2 px} & \multicolumn{2}{c|}{\textgreater3 px} & \multicolumn{2}{c|}{\textgreater 5 px} & \multicolumn{2}{c|}{Mean Error} & Runtime \\
& Non-Occ & All & Non-Occ & All & Non-Occ & All & Non-Occ & All & (s) \\ \hline \hline
SPS-st \cite{yamaguchi2014efficient}  		& 4.98 & 6.28 & 3.39 & 4.41 & 2.33 & 3.00 & 0.9 px & 1.0 px & 2 \\
Deep Embed \cite{chen2015deep} 				& 5.05 & 6.47 & 3.10 & 4.24 & 1.92 & 2.68 & 0.9 px & 1.1 px & 3 \\
Content-CNN \cite{luo2016efficient}  		& 4.98 & 6.51 & 3.07 & 4.29 & 2.03 & 2.82 & 0.8 px & 1.0 px & \textbf{0.7} \\ 
MC-CNN \cite{zbontar2016stereo} 			& 3.90 & 5.45 & 2.43 & 3.63 & 1.64 & 2.39 & 0.7 px & 0.9 px & 67 \\
PBCP \cite{Seki2016BMVC} 					& 3.62 & 5.01 & 2.36 & 3.45 & 1.62 & 2.32 & 0.7 px & 0.9 px & 68 \\
Displets v2 \cite{guney2015displets} 		& 3.43 & 4.46 & 2.37 & 3.09 & 1.72 & 2.17 & 0.7 px & 0.8 px & 265 \\ \hline
GC-Net (this work)           				& \textbf{2.71} & \textbf{3.46} & \textbf{1.77} & \textbf{2.30} & \textbf{1.12} & \textbf{1.46} & \textbf{0.6 px} & \textbf{0.7 px} & 0.9 \\
\end{tabular}
	\caption{\textbf{KITTI 2012 test set results} \cite{Geiger2012CVPR}. This benchmark contains 194 train and 195 test gray-scale image pairs.}
	\label{tbl:kitti2012}
\end{subtable}

\vspace{5 mm}

\begin{subtable}[t]{\linewidth}
\centering
\begin{tabular}{l|ccc|ccc|c} \hline
                   & \multicolumn{3}{c|}{All Pixels} & \multicolumn{3}{c|}{Non-Occluded Pixels} & Runtime \\
                   & D1-bg   & D1-fg   & D1-all  & D1-bg   & D1-fg   & D1-all  & (s)     \\ \hline \hline
MBM \cite{einecke2015multi}			& 4.69 & 13.05 & 6.08 & 4.33 & 12.12 & 5.61 & 0.13 \\
ELAS \cite{geiger2010efficient} 	& 7.86 & 19.04 & 9.72 & 6.88 & 17.73 & 8.67 & 0.3 \\
Content-CNN \cite{luo2016efficient} & 3.73 & 8.58  & 4.54 & 3.32 & 7.44  & 4.00 & 1.0 \\
DispNetC \cite{Mayer2015} 			& 4.32 & \textbf{4.41}  & 4.34 & 4.11 & \textbf{3.72}  & 4.05 & \textbf{0.06}\\
MC-CNN \cite{zbontar2016stereo}     & 2.89 & 8.88  & 3.89 & 2.48 & 7.64  & 3.33 & 67  \\
PBCP \cite{Seki2016BMVC} 			& 2.58 & 8.74  & 3.61 & 2.27 & 7.71  & 3.17 & 68  \\
Displets v2 \cite{guney2015displets}& 3.00& 5.56  & 3.43 & 2.73 & 4.95  & 3.09 & 265 \\ \hline
GC-Net (this work)					& \textbf{2.21} & 6.16  & \textbf{2.87} & \textbf{2.02} & 5.58  & \textbf{2.61} & 0.9 \\      
\end{tabular}
	\caption{\textbf{KITTI 2015 test set results} \cite{Menze2015CVPR}. This benchmark contains 200 training and 200 test color image pairs. The qualifier `bg' refers to background pixels which contain static elements, `fg' refers to dynamic object pixels, while `all' is all pixels (fg+bg). The results show the percentage of pixels which have greater than three pixels or 5\% disparity error from all 200 test images.}
	\label{tbl:kitti2015}
\end{subtable}
	\caption{Comparison to other stereo methods on the test set of \textbf{KITTI 2012 and 2015 benchmarks} \cite{Geiger2012CVPR,Menze2015CVPR}. Our method sets a new state-of-the-art on these two competitive benchmarks, out performing all other approaches.}
	\label{tbl:kitti_test}
\end{table*}

\begin{figure}[t]
\centering
\includegraphics[width=\linewidth]{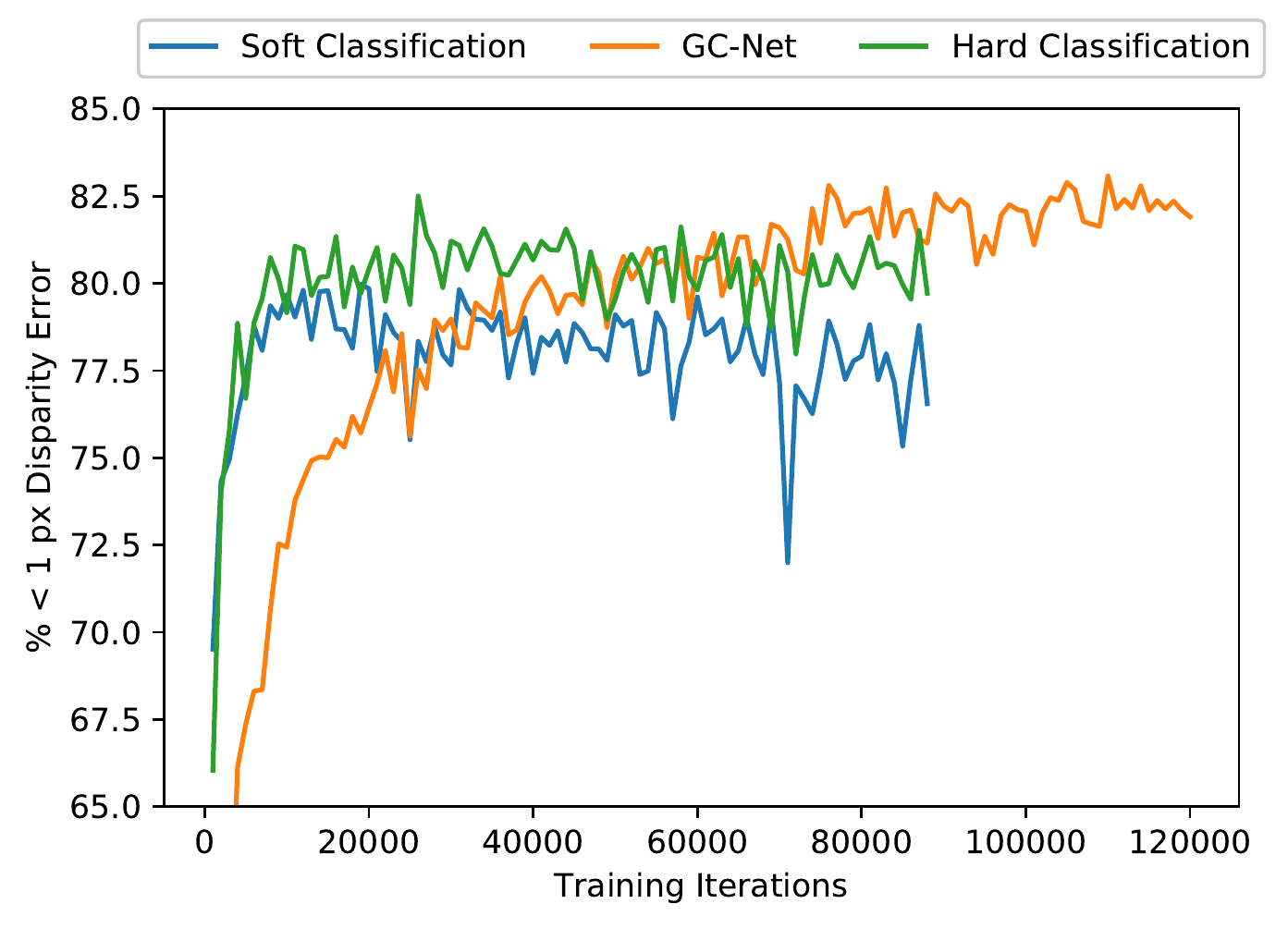}
\caption{\textbf{Validation error} (percentage of disparities with error less than 1 px) during training with the Scene Flow dataset. Classification loss trains faster, however using a regression loss results in better performance.}
\label{fig:training_err}
\end{figure}

\fig{training_err} plots validation error during training for each of the networks compared in this section. We observe that the classification loss converges faster, however the regression loss performs best overall.

\begin{figure*}[p]
	\begin{center}
            \vspace{-5 mm}
    		\begin{subfigure}[t]{\linewidth}
            \centering
			\includegraphics[width=0.31\linewidth]{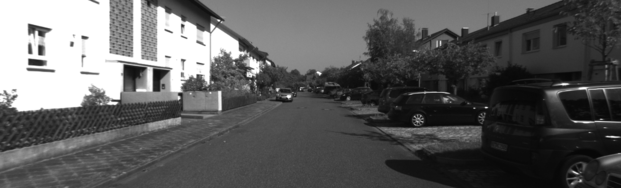}
			\includegraphics[width=0.31\linewidth]{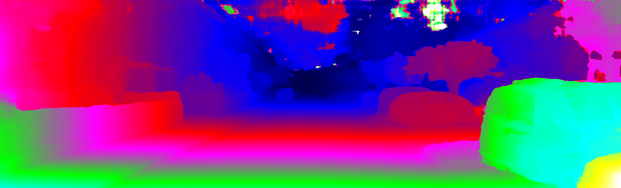}
			\includegraphics[width=0.31\linewidth]{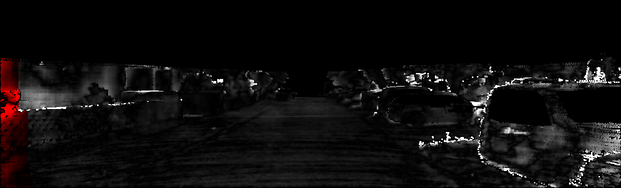}
            \vspace{1 mm}
			\includegraphics[width=0.31\linewidth]{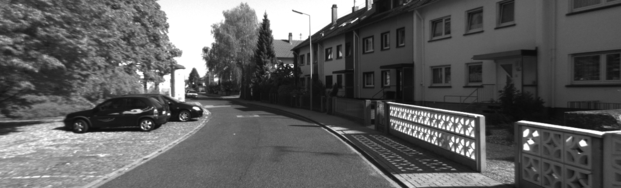}
			\includegraphics[width=0.31\linewidth]{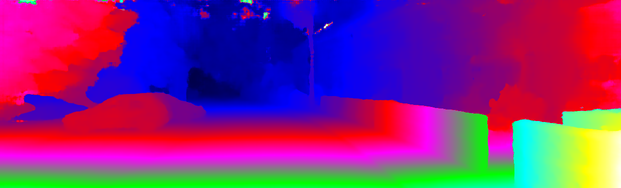}
			\includegraphics[width=0.31\linewidth]{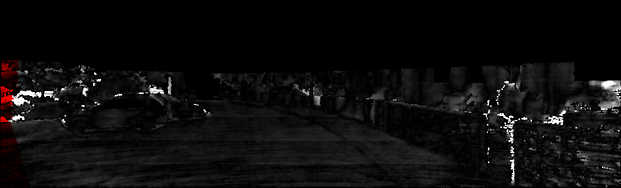}
            \vspace{1 mm}
			\includegraphics[width=0.31\linewidth]{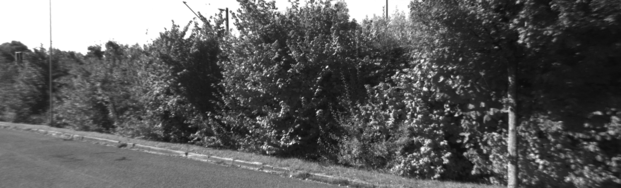}
			\includegraphics[width=0.31\linewidth]{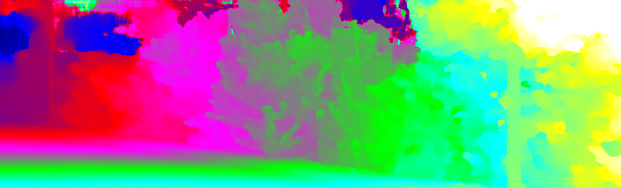}
			\includegraphics[width=0.31\linewidth]{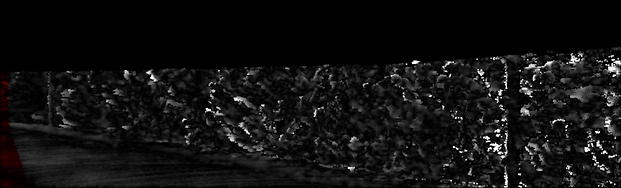}
	        \caption{KITTI 2012 test data qualitative results. From left: left stereo input image, disparity prediction, error map.}
            \vspace{2 mm}
		\end{subfigure}
    		\begin{subfigure}[t]{\linewidth}
            \centering
			\includegraphics[width=0.31\linewidth]{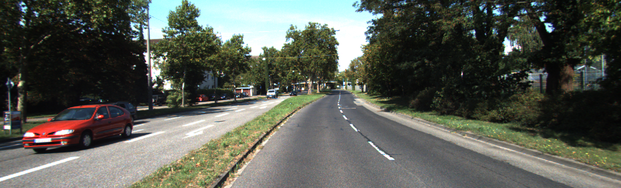}
			\includegraphics[width=0.31\linewidth]{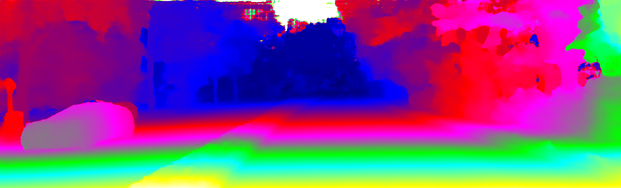}
			\includegraphics[width=0.31\linewidth]{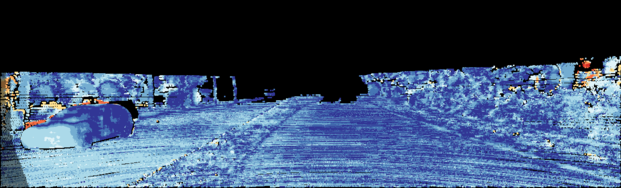}
            \vspace{1 mm}
			\includegraphics[width=0.31\linewidth]{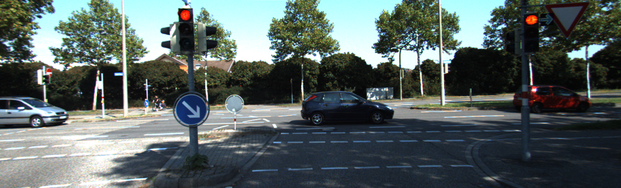}
			\includegraphics[width=0.31\linewidth]{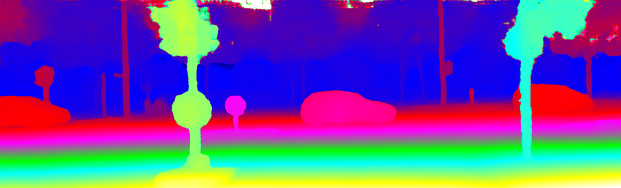}
			\includegraphics[width=0.31\linewidth]{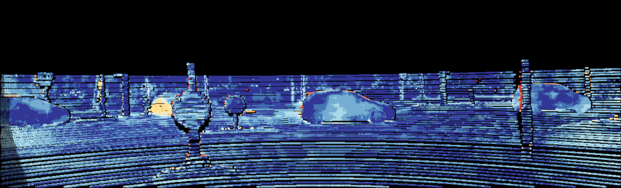}
            \vspace{1 mm}
			\includegraphics[width=0.31\linewidth]{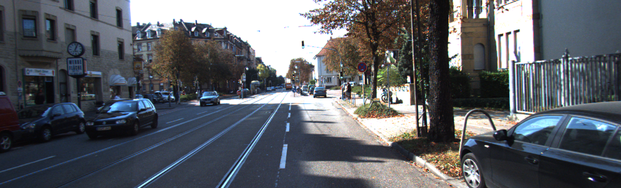}
			\includegraphics[width=0.31\linewidth]{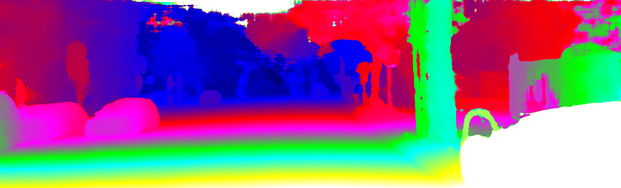}
			\includegraphics[width=0.31\linewidth]{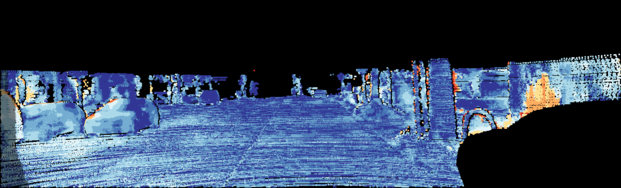}
	        \caption{KITTI 2015 test data qualitative results. From left: left stereo input image, disparity prediction, error map.}
            \vspace{2 mm}
		\end{subfigure}
    		\begin{subfigure}[t]{\linewidth}
            \centering
			\includegraphics[width=0.31\linewidth]{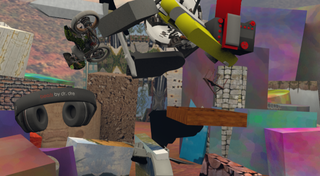}
			\includegraphics[width=0.31\linewidth]{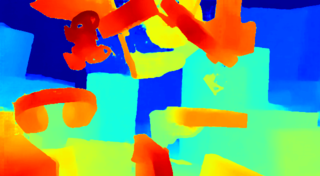}
			\includegraphics[width=0.31\linewidth]{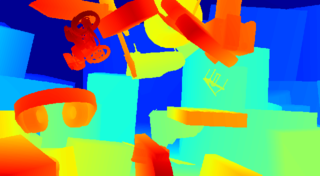}
            \vspace{1 mm}
			\includegraphics[width=0.31\linewidth]{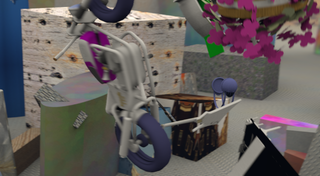}
			\includegraphics[width=0.31\linewidth]{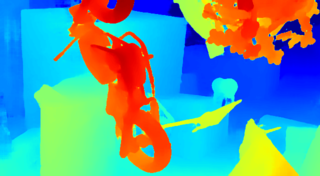}
			\includegraphics[width=0.31\linewidth]{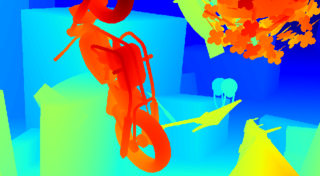}
            \vspace{1 mm}
			\includegraphics[width=0.31\linewidth]{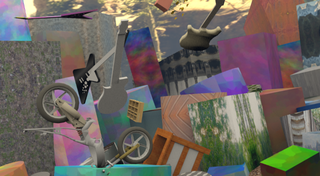}
			\includegraphics[width=0.31\linewidth]{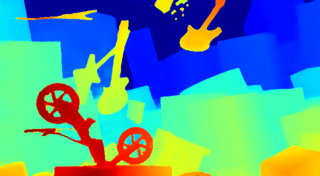}
			\includegraphics[width=0.31\linewidth]{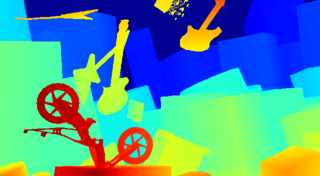}
	        \caption{Scene Flow test set qualitative results. From left: left stereo input image, disparity prediction, ground truth.}
		\end{subfigure}
	\end{center}
	\caption{\textbf{Qualitative results.} By learning to incorporate wider context our method is often able to handle challenging scenarios, such as reflective, thin or texture-less surfaces. By explicitly learning geometry in a cost volume, our method produces sharp results and can also handle large occlusions.}
	\label{fig:results_qualitative}
\end{figure*}

\begin{figure}[t]
	\begin{center}
    		\begin{subfigure}[t]{\linewidth}
            \centering
            \includegraphics[width=0.32\linewidth,trim={0 0 0 1cm},clip]{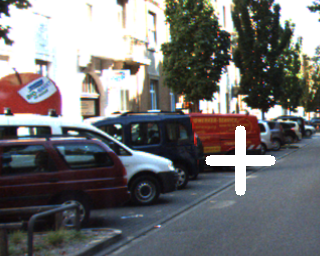}
            \includegraphics[width=0.32\linewidth,trim={0 0 0 1cm},clip]{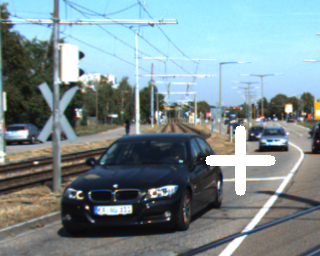}
            \includegraphics[width=0.32\linewidth,trim={0 0 0 1cm},clip]{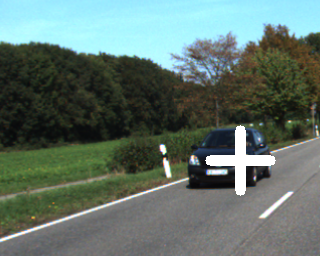}
	        \caption{Left stereo input image}
		\end{subfigure}
    		\begin{subfigure}[t]{\linewidth}
            \centering
            \includegraphics[width=0.32\linewidth,trim={0 0 0 1cm},clip]{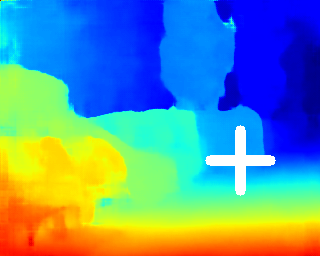}
            \includegraphics[width=0.32\linewidth,trim={0 0 0 1cm},clip]{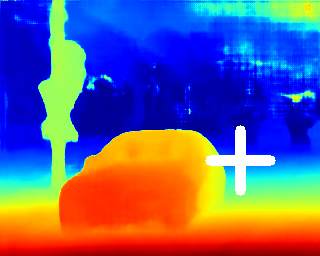}
            \includegraphics[width=0.32\linewidth,trim={0 0 0 1cm},clip]{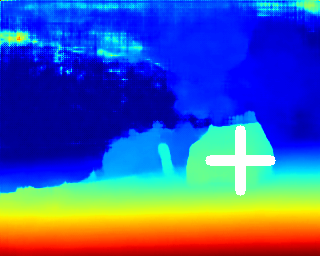}
	        \caption{Predicted disparity map}
		\end{subfigure}
    		\begin{subfigure}[t]{\linewidth}
            \centering
            \includegraphics[width=0.32\linewidth,trim={0 0 0 1cm},clip]{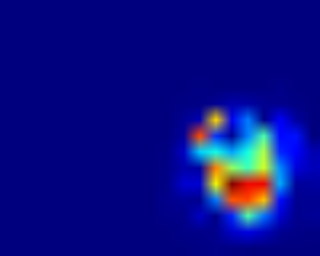}
            \includegraphics[width=0.32\linewidth,trim={0 0 0 1cm},clip]{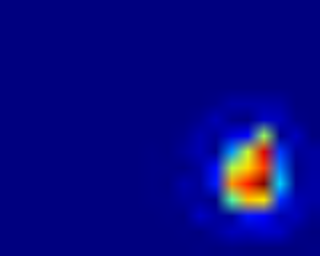}
            \includegraphics[width=0.32\linewidth,trim={0 0 0 1cm},clip]{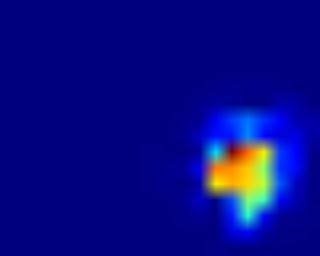}
	        \caption{Saliency map (red = stronger saliency)}
		\end{subfigure}
    		\begin{subfigure}[t]{\linewidth}
            \centering
            \includegraphics[width=0.32\linewidth,trim={0 0 0 1cm},clip]{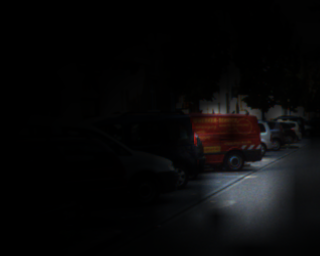}
            \includegraphics[width=0.32\linewidth,trim={0 0 0 1cm},clip]{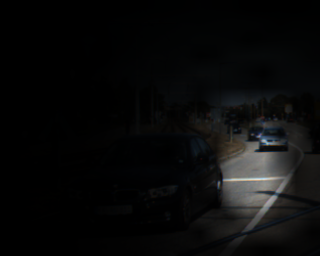}
            \includegraphics[width=0.32\linewidth,trim={0 0 0 1cm},clip]{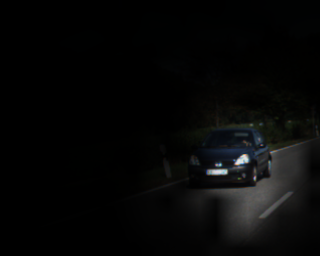}
	        \caption{What the network sees (input attenuated by saliency)}
		\end{subfigure}
	\end{center}
    \vspace{-2mm}
	\caption{\textbf{Saliency map visualization} which shows the model's effective receptive field for a selected output pixel (indicated by the white cross). This shows that our architecture is able to learn to regress stereo disparity with a large field of view and significant contextual knowledge of the scene, beyond the local geometry and appearance. For example, in the example on the right we observe that the model considers contextual information from the vehicle and surrounding road surface to estimate disparity.}
	\label{fig:saliency}
\end{figure}

\subsection{KITTI Benchmark}
\label{sec:kitti_results}

In \tbl{kitti_test} we evaluate the performance of our model on the KITTI 2012 and 2015 stereo datasets \cite{Geiger2012CVPR,Menze2015CVPR}. These consist of challenging and varied road scene imagery collected from a test vehicle. Ground truth depth maps for training and evaluation are obtained from LIDAR data. KITTI is a prominent dataset for benchmarking stereo algorithms. The downside is that it only contains $200$ training images, which handicaps learning algorithms. for this reason, we pre-train our model on the large synthetic dataset, Scene Flow \cite{MIFDB16}. This helps to prevent our model from over-fitting the very small KITTI training dataset. We hold out 40 image pairs as our validation set.

\tbl{kitti2012} and \ref{tbl:kitti2015} compare our method, GC-Net (\underline{G}eometry and \underline{C}ontext \underline{Net}work), to other approaches on the KITTI 2012 and 2015 datasets, respectively\footnote{Full leaderboard:~\url{www.cvlibs.net/datasets/kitti/}}. Our method achieves state of the art results for both KITTI benchmarks, by a notable margin. We improve on state-of-the-art by 9\% and 22\% for KITTI 2015 and 2012 respectively. Our method is also notably faster than most competing approaches which often require expensive post-processing. In \fig{results_qualitative} we show qualitative results of our method on KITTI 2012, KITTI 2015 and Scene Flow.

Our approach outperforms previous deep learning patch based methods \cite{zbontar2015computing,luo2016efficient} which produce noisy unary potentials and are unable to predict with sub-pixel accuracy. For this reason, these algorithms do not use end-to-end learning and typically post-process the unary output with SGM regularization \cite{einecke2015multi} to produce the final disparity maps.

The closest method to our architecture is DispNetC \cite{Mayer2015}, which is an end-to-end regression network pre-trained on SceneFlow. However, our method outperforms this architecture by a notable margin for \textit{all} test pixels. DispNetC uses a 1-D correlation layer along the disparity line as an approximation to the stereo cost volume. In contrast, our architecture more explicitly leverages geometry by formulating a full cost volume by using 3-D convolutions and a soft argmin layer, resulting in an improvement in performance.




\subsection{Model Saliency}
\label{sec:saliency}

In this section we present evidence which shows our model can reason about local geometry using wider contextual information. In \fig{saliency} we show some examples of the model's saliency with respect to a predicted pixel's disparity. Saliency maps \cite{simonyan2013deep} shows the sensitivity of the output with respect to each input pixel. We use the method from \cite{zeiler2014visualizing} which plots the predicted disparity as a function of systematically occluding the input images. We offset the occlusion in each stereo image by the point's disparity.

These results show that the disparity prediction for a given point is dependent on a wide contextual field of view. For example, the disparity on the front of the car depends on the input pixels of the car and the road surface below. This demonstrates that our model is able to reason about wider context, rather than simply $9\times9$ local patches like previous deep learning patch-similarity stereo methods \cite{zbontar2016stereo,luo2016efficient}.

\section{Conclusions}

We propose a novel end-to-end deep learning architecture for stereo vision. It is able to learn to regress disparity without any additional post-processing or regularization. We demonstrate the efficacy of our method on the KITTI dataset, setting a new state-of-the-art benchmark.

We show how to efficiently learn context in the disparity cost volume using 3-D convolutions. We show how to formulate it as a regression model using a soft argmin layer. This allows us to learn disparity as a regression problem, rather than classification, improving performance and enabling sub-pixel accuracy. We demonstrate that our model learns to incorporate wider contextual information.

For future work we are interested in exploring a more explicit representation of semantics to improve our disparity estimation, and reasoning under uncertainty with Bayesian convolutional neural networks.

{\footnotesize
\bibliographystyle{ieee}
\bibliography{deep_stereo}

\begin{thebibliography}{10}\itemsep=-1pt

\bibitem{tensorflow2015-whitepaper}
M.~Abadi, A.~Agarwal, P.~Barham, E.~Brevdo, Z.~Chen, C.~Citro, G.~S. Corrado,
  A.~Davis, J.~Dean, M.~Devin, S.~Ghemawat, I.~Goodfellow, A.~Harp, G.~Irving,
  M.~Isard, Y.~Jia, R.~Jozefowicz, L.~Kaiser, M.~Kudlur, J.~Levenberg,
  D.~Man\'{e}, R.~Monga, S.~Moore, D.~Murray, C.~Olah, M.~Schuster, J.~Shlens,
  B.~Steiner, I.~Sutskever, K.~Talwar, P.~Tucker, V.~Vanhoucke, V.~Vasudevan,
  F.~Vi\'{e}gas, O.~Vinyals, P.~Warden, M.~Wattenberg, M.~Wicke, Y.~Yu, and
  X.~Zheng.
\newblock {TensorFlow}: Large-scale machine learning on heterogeneous systems,
  2015.
\newblock Software available from tensorflow.org.

\bibitem{achtelik2009stereo}
M.~Achtelik, A.~Bachrach, R.~He, S.~Prentice, and N.~Roy.
\newblock Stereo vision and laser odometry for autonomous helicopters in
  gps-denied indoor environments.
\newblock In {\em SPIE Defense, security, and sensing}, pages 733219--733219.
  International Society for Optics and Photonics, 2009.

\bibitem{badrinarayanan2015segnet}
V.~Badrinarayanan, A.~Kendall, and R.~Cipolla.
\newblock Segnet: A deep convolutional encoder-decoder architecture for image
  segmentation.
\newblock {\em arXiv preprint arXiv:1511.00561}, 2015.

\bibitem{bahdanau2014neural}
D.~Bahdanau, K.~Cho, and Y.~Bengio.
\newblock Neural machine translation by jointly learning to align and
  translate.
\newblock In {\em ICLR 2015}, 2014.

\bibitem{Barnard1982}
S.~T. Barnard and M.~A. Fischler.
\newblock {Computational stereo}.
\newblock {\em ACM Computing Surveys}, 14(4):553--572, 1982.

\bibitem{Bleyer2011}
M.~Bleyer, C.~Rhemann, and C.~Rother.
\newblock {PatchMatch Stereo-Stereo Matching with Slanted Support Windows.}
\newblock {\em Bmvc}, i(1):14.1--14.11, 2011.

\bibitem{Calonder2010}
M.~Calonder, V.~Lepetit, and C.~Strecha.
\newblock {BRIEF : Binary Robust Independent Elementary Features}.
\newblock In {\em European Conference on Computer Vision (ECCV)}, 2010.

\bibitem{chen2015deep}
Z.~Chen, X.~Sun, L.~Wang, Y.~Yu, and C.~Huang.
\newblock A deep visual correspondence embedding model for stereo matching
  costs.
\newblock In {\em Proceedings of the IEEE International Conference on Computer
  Vision}, pages 972--980, 2015.

\bibitem{Chen2016}
Z.~Chen, X.~Sun, L.~Wang, Y.~Yu, and C.~Huang.
\newblock {A deep visual correspondence embedding model for stereo matching
  costs}.
\newblock In {\em Proceedings of the IEEE International Conference on Computer
  Vision}, pages 972--980, 2016.

\bibitem{Eigen2014a}
D.~Eigen, C.~Puhrsch, and R.~Fergus.
\newblock {Depth map prediction from a single image using a multi-scale deep
  network}.
\newblock {\em Nips}, pages 1--9, 2014.

\bibitem{einecke2015multi}
N.~Einecke and J.~Eggert.
\newblock A multi-block-matching approach for stereo.
\newblock In {\em 2015 IEEE Intelligent Vehicles Symposium (IV)}, pages
  585--592. IEEE, 2015.

\bibitem{Flynn2016}
J.~Flynn, I.~Neulander, J.~Philbin, and N.~Snavely.
\newblock {DeepStereo: Learning to Predict New Views from the World's Imagery}.
\newblock {\em CVPR}, 2016.

\bibitem{Garg2016}
R.~Garg, V.~{Kumar BG}, and I.~Reid.
\newblock {Unsupervised CNN for Single View Depth Estimation: Geometry to the
  Rescue}.
\newblock {\em ECCV}, pages 1--16, 2016.

\bibitem{Geiger2012CVPR}
A.~Geiger, P.~Lenz, and R.~Urtasun.
\newblock Are we ready for autonomous driving? the kitti vision benchmark
  suite.
\newblock In {\em Conference on Computer Vision and Pattern Recognition
  (CVPR)}, 2012.

\bibitem{geiger2010efficient}
A.~Geiger, M.~Roser, and R.~Urtasun.
\newblock Efficient large-scale stereo matching.
\newblock In {\em Asian conference on computer vision}, pages 25--38. Springer,
  2010.

\bibitem{Geiger2010}
A.~Geiger, M.~Roser, and R.~Urtasun.
\newblock {Efficient Large-Scale Stereo Matching}.
\newblock {\em Computer Vision – ACCV 2010}, (1):25--38, 2010.

\bibitem{girshick2014rich}
R.~Girshick, J.~Donahue, T.~Darrell, and J.~Malik.
\newblock Rich feature hierarchies for accurate object detection and semantic
  segmentation.
\newblock In {\em Proceedings of the IEEE conference on computer vision and
  pattern recognition}, pages 580--587, 2014.

\bibitem{guney2015displets}
F.~Guney and A.~Geiger.
\newblock Displets: Resolving stereo ambiguities using object knowledge.
\newblock In {\em Proceedings of the IEEE Conference on Computer Vision and
  Pattern Recognition}, pages 4165--4175, 2015.

\bibitem{Haeusler2013a}
R.~Haeusler, R.~Nair, and D.~Kondermann.
\newblock {Ensemble Learning for Confidence Measures in Stereo Vision}.
\newblock {\em Computer Vision and Pattern Recognition (CVPR), 2013 IEEE
  Conference on}, pages 305--312, 2013.

\bibitem{hartley2003multiple}
R.~Hartley and A.~Zisserman.
\newblock {\em Multiple view geometry in computer vision}.
\newblock Cambridge university press, 2003.

\bibitem{he2015deep}
K.~He, X.~Zhang, S.~Ren, and J.~Sun.
\newblock Deep residual learning for image recognition.
\newblock In {\em In Proc. IEEE Conf. on Computer Vision and Pattern
  Recognition}, 2016.

\bibitem{Heise2015}
P.~Heise, B.~Jensen, S.~Klose, and A.~Knoll.
\newblock {Fast Dense Stereo Correspondences by Binary Locality Sensitive
  Hashing}.
\newblock {\em ICRA}, pages 1--6, 2015.

\bibitem{hirschmuller2005accurate}
H.~Hirschmuller.
\newblock Accurate and efficient stereo processing by semi-global matching and
  mutual information.
\newblock In {\em 2005 IEEE Computer Society Conference on Computer Vision and
  Pattern Recognition (CVPR'05)}, volume~2, pages 807--814. IEEE, 2005.

\bibitem{Hirschmuller2008}
H.~Hirschm{\"{u}}ller.
\newblock {Stereo processing by semiglobal matching and mutual information}.
\newblock {\em IEEE Transactions on Pattern Analysis and Machine Intelligence},
  30(2):328--341, 2008.

\bibitem{Hirschmuller2007}
H.~Hirschm{\"{u}}ller and D.~Scharstein.
\newblock {Evaluation of Cost Functions for Stereo Matching}.
\newblock In {\em 2007 IEEE Conference on Computer Vision and Pattern
  Recognition}, 2007.

\bibitem{Klaus2006}
A.~Klaus, M.~Sormann, and K.~Karner.
\newblock {Segment-based stereo matching using belief propagation and a
  self-adapting dissimilarity measure}.
\newblock {\em Proceedings - International Conference on Pattern Recognition},
  3:15--18, 2006.

\bibitem{Kolmogorov2001}
V.~Kolmogorov and R.~Zabih.
\newblock {Computing visual correspondences with occlusions using graph cuts}.
\newblock In {\em International Conference on Computer Vision (ICCV)}, 2001.

\bibitem{krizhevsky2012imagenet}
A.~Krizhevsky, I.~Sutskever, and G.~E. Hinton.
\newblock Imagenet classification with deep convolutional neural networks.
\newblock In {\em Advances in neural information processing systems}, pages
  1097--1105, 2012.

\bibitem{Li2008}
Y.~Li and D.~P. Huttenlocher.
\newblock {Learning for stereo vision using the structured support vector
  machine}.
\newblock In {\em 2008 IEEE Conference on Computer Vision and Pattern
  Recognition}, 2008.

\bibitem{Liu2015}
F.~Liu, C.~Shen, G.~Lin, and I.~Reid.
\newblock {Learning Depth from Single Monocular Images Using Deep Convolutional
  Neural Fields}.
\newblock {\em Pattern Analysis and Machine Intelligence}, page~15, 2015.

\bibitem{long2015fully}
J.~Long, E.~Shelhamer, and T.~Darrell.
\newblock Fully convolutional networks for semantic segmentation.
\newblock In {\em Proceedings of the IEEE Conference on Computer Vision and
  Pattern Recognition}, pages 3431--3440, 2015.

\bibitem{luo2016efficient}
W.~Luo, A.~G. Schwing, and R.~Urtasun.
\newblock Efficient deep learning for stereo matching.
\newblock In {\em Proceedings of the IEEE Conference on Computer Vision and
  Pattern Recognition}, pages 5695--5703, 2016.

\bibitem{Luo2016}
W.~Luo, A.~G. Schwing, and R.~Urtasun.
\newblock {Efficient Deep Learning for Stereo Matching}.
\newblock {\em CVPR}, 2016.

\bibitem{Mayer2015}
N.~Mayer, E.~Ilg, P.~H{\"{a}}usser, P.~Fischer, D.~Cremers, A.~Dosovitskiy, and
  T.~Brox.
\newblock {A Large Dataset to Train Convolutional Networks for Disparity,
  Optical Flow, and Scene Flow Estimation}.
\newblock {\em CoRR}, abs/1510.0(2002), 2015.

\bibitem{Menze2015CVPR}
M.~Menze and A.~Geiger.
\newblock Object scene flow for autonomous vehicles.
\newblock In {\em Conference on Computer Vision and Pattern Recognition
  (CVPR)}, 2015.

\bibitem{MIFDB16}
N.Mayer, E.Ilg, P.H{\"a}usser, P.Fischer, D.Cremers, A.Dosovitskiy, and T.Brox.
\newblock A large dataset to train convolutional networks for disparity,
  optical flow, and scene flow estimation.
\newblock In {\em IEEE International Conference on Computer Vision and Pattern
  Recognition (CVPR)}, 2016.
\newblock arXiv:1512.02134.

\bibitem{Park2015}
M.~G. Park and K.~J. Yoon.
\newblock {Leveraging stereo matching with learning-based confidence measures}.
\newblock {\em Proceedings of the IEEE Computer Society Conference on Computer
  Vision and Pattern Recognition}, 07-12-June:101--109, 2015.

\bibitem{Scharstein2007}
D.~Scharstein and C.~Pal.
\newblock {Learning conditional random fields for stereo}.
\newblock {\em Proceedings of the IEEE Computer Society Conference on Computer
  Vision and Pattern Recognition}, 2007.

\bibitem{Scharstein2002}
D.~Scharstein and R.~Szeliski.
\newblock {A Taxonomy and Evaluation of Dense Two-Frame Stereo Correspondence
  Algorithms}.
\newblock {\em International Journal of Computer Vision}, 47(1):7--42, 2002.

\bibitem{Seki2016BMVC}
A.~Seki and M.~Pollefeys.
\newblock Patch based confidence prediction for dense disparity map.
\newblock In {\em British Machine Vision Conference (BMVC)}, 2016.

\bibitem{simonyan2013deep}
K.~Simonyan, A.~Vedaldi, and A.~Zisserman.
\newblock Deep inside convolutional networks: Visualising image classification
  models and saliency maps.
\newblock {\em arXiv preprint arXiv:1312.6034}, 2013.

\bibitem{tieleman2012lecture}
T.~Tieleman and G.~Hinton.
\newblock Lecture 6.5-rmsprop: Divide the gradient by a running average of its
  recent magnitude.
\newblock {\em COURSERA: Neural networks for machine learning}, 4(2), 2012.

\bibitem{Tombari2008}
F.~Tombari, S.~Mattoccia, L.~D. Stefano, and E.~Addimanda.
\newblock {Classification and evaluation of cost aggregation methods for stereo
  correspondence}.
\newblock {\em 26th IEEE Conference on Computer Vision and Pattern Recognition,
  CVPR}, 2008.

\bibitem{yamaguchi2014efficient}
K.~Yamaguchi, D.~McAllester, and R.~Urtasun.
\newblock Efficient joint segmentation, occlusion labeling, stereo and flow
  estimation.
\newblock In {\em European Conference on Computer Vision}, pages 756--771.
  Springer, 2014.

\bibitem{Zabih1994}
R.~Zabih and J.~Woodfill.
\newblock {Non-parametric Local Transforms for Computing Visual
  Correspondence}.
\newblock {\em In Proceedings of European Conference on Computer Vision},
  (May):151--158, 1994.

\bibitem{Zagoruyko2015}
S.~Zagoruyko and N.~Komodakis.
\newblock {Learning to compare image patches via convolutional neural
  networks}.
\newblock {\em Proceedings of the IEEE Computer Society Conference on Computer
  Vision and Pattern Recognition}, 07-12-June(i):4353--4361, 2015.

\bibitem{Zbontar2015a}
J.~{\v{Z}}bontar and Y.~{Le Cun}.
\newblock {Computing the stereo matching cost with a convolutional neural
  network}.
\newblock {\em Proceedings of the IEEE Computer Society Conference on Computer
  Vision and Pattern Recognition}, 07-12-June(1):1592--1599, 2015.

\bibitem{zbontar2015computing}
J.~Zbontar and Y.~LeCun.
\newblock Computing the stereo matching cost with a convolutional neural
  network.
\newblock In {\em Proceedings of the IEEE Conference on Computer Vision and
  Pattern Recognition}, pages 1592--1599, 2015.

\bibitem{Zbontar2015}
J.~{\v{Z}}bontar and Y.~LeCun.
\newblock {Stereo Matching by Training a Convolutional Neural Network to
  Compare Image Patches}.
\newblock {\em CoRR}, abs/1510.0(2002), 2015.

\bibitem{zbontar2016stereo}
J.~Zbontar and Y.~LeCun.
\newblock Stereo matching by training a convolutional neural network to compare
  image patches.
\newblock {\em Journal of Machine Learning Research}, 17:1--32, 2016.

\bibitem{zeiler2014visualizing}
M.~D. Zeiler and R.~Fergus.
\newblock Visualizing and understanding convolutional networks.
\newblock In {\em European conference on computer vision}, pages 818--833.
  Springer, 2014.

\bibitem{Zhang2007}
L.~Zhang and S.~M. Seitz.
\newblock {Estimating optimal parameters for {\{}MRF{\}} stereo from a single
  image pair}.
\newblock {\em IEEE Transactions on Pattern Analysis and Machine Intelligence},
  29(2):331--342, 2007.

\end{thebibliography}
}

\end{document}